\documentclass{tlp_modificata}
\usepackage{latexsym}
\usepackage{color}
\usepackage{oldlfont,rawfonts}
\usepackage{graphicx}
\usepackage{verbatim}
\usepackage{courier}

\usepackage{amssymb}
\usepackage{amsmath}

\usepackage{todonotes}
\usepackage{url}
\usepackage{booktabs}
\usepackage{multirow}
\usepackage{pgfplots}\usetikzlibrary{plotmarks}\pgfplotsset{compat=1.14}

\usepackage{caption}
    \usepackage[switch]{lineno}
    \usepackage{etoolbox} 
    
    \newcommand*\linenomathpatch[1]{%
      \cspreto{#1}{\linenomath}%
      \cspreto{#1*}{\linenomath}%
      \csappto{end#1}{\endlinenomath}%
      \csappto{end#1*}{\endlinenomath}%
    }
    
    \linenomathpatch{equation}
    \linenomathpatch{gather}
    \linenomathpatch{multline}
    \linenomathpatch{align}
    \linenomathpatch{alignat}
    \linenomathpatch{flalign}

\usepackage{listings}
\makeatletter
\lst@Key{countblanklines}{true}[t]{\lstKV@SetIf{#1}\lst@ifcountblanklines}
\lst@AddToHook{OnEmptyLine}{%
    \vspace{-0.7\baselineskip}%
    \lst@ifnumberblanklines\else%
    \lst@ifcountblanklines\else%
    \advance\c@lstnumber-\@ne\relax%
    \fi%
    \fi}
\makeatother
\newcommand {\ent} {\mathrel{{\scriptstyle\mid\!\sim}}}

\newcommand{\tip}{{\bf T}}

\newcommand{\alc}{\mathcal{ALC}}
\newcommand{\lc}{\mathcal{LC}}

\newcommand{\be}{\begin{enumerate}}
\newcommand{\ee}{\end{enumerate}}

\newcommand{\hide}[1]{}

\def \cases{\left \{\begin{array}{l}}
\def \endcases{\end{array}\right .}

\newcommand {\ri} {\rightarrow}

\newcommand {\bes} {\begin{description}}
\newcommand{\ens} {\end{description}}
\newcommand {\la} {\langle}
\newcommand {\ra} {\rangle}

\newcommand {\beq} {\begin{quote}}
\newcommand {\enq} {\end{quote}}
\newcommand {\bit} {\begin{itemize}}
\newcommand {\enit} {\end{itemize}}

\newtheorem{theorem}{Theorem}
\newtheorem{lemma}{Lemma}

\newtheorem{proposition}{Proposition}
\newtheorem{definition}{Definition}

\begin{document}

\bibliographystyle{plain}

\lefttitle{M. Alviano, L. Giordano and D. Theseider Dupr{\'e} }


 \title[Theory and Practice of Logic Programming]
 {
Complexity and scalability of defeasible reasoning \\
in many-valued weighted knowledge bases \\
 with typicality
}


\begin{authgrp}
\author{Mario Alviano}
\affiliation{Universit\`a della Calabria, Italy}
\author{Laura Giordano, Daniele Theseider Dupr{\'e} }
\affiliation{Universit\`a del Piemonte Orientale, Italy}
\end{authgrp}


\maketitle
 
\begin{abstract}
Weighted knowledge bases for description logics with typicality under a ``concept-wise'' multi-preferential semantics provide a logical interpretation of MultiLayer Perceptrons.
In this context, Answer Set Programming (ASP) has been shown to be suitable for addressing defeasible reasoning in the finitely many-valued case, providing a $\Pi^p_2$ upper bound on the complexity of the problem, nonetheless leaving unknown the exact complexity and only providing a proof-of-concept implementation.
This paper fulfils the lack by providing a $\sc{P^{NP[log]}}$-completeness result and new ASP encodings that deal with 
weighted knowledge bases with large search spaces.
\end{abstract}


\section{Introduction}


Description logics (DLs) are widely used for knowledge representation (KR), often to verify and discover properties of individuals in a concept by means of DLs inference services \citep{handbook,DBLP:books/crc/Hitzler2010}.
Many properties of real world concepts, however, are {\em defeasible}, that is, they are not universally true, but have exceptions, and actually hold only for some \emph{typical} individuals in the concept.
For example, horses are usually tall, but \emph{atypical} horses not being tall exist.
This has led to a line of research which deals with {\em defeasible DLs} 
\citep{BritzKR08,FI09,casinistraccia2010}.
Specifically, to represent the defeasible properties of a concept,  DLs can be extended with a \emph{typicality operator} $\tip$ that is applied to concepts to obtain \emph{typicality inclusions} of the form $\tip(C) \sqsubseteq D$ \citep{FI09}.
Intuitively, $\tip(C) \sqsubseteq D$ means that the typical individuals in the concept $C$ also belong to concept $D$ (that, {\em normally} C's are D's), and corresponds to a \emph{conditional implication} $C \ent D$ in KLM preferential logics \citep{KrausLehmannMagidor:90, whatdoes}.
A (conditional) knowledge base (KB) comprising typicality inclusions enables \emph{defeasible reasoning}, as in fact properties holding for typical individuals in $C$ are not necessarily enforced on all individuals in $C$.

Some control on the strength of the applicability of typicality inclusions (which, otherwise, depends on specificity) is obtained by assigning them a rank, that is, a natural number as large as strong is the expressed property.
The resulting \emph{ranked DL KBs} --- reminiscent of ranked KBs by Brewka (\citeyear{Brewka04}) --- are interpreted according to a concept-wise {\em multi-preferential} semantics, that is, by associating a preference relation to single concepts to identify \emph{the most typical} individuals in a concept \citep{TPLP2020}. 
A more fine-grained control is obtained by assigning weights to typicality inclusions, hence obtaining \emph{weighted DL KBs} \citep{JELIA2021}.
In fact, weighing typicality inclusions with positive and negative real numbers allow for representing their plausibility or implausibility.
A concrete application of the extended concept-wise multi-preferential semantics is represented by the \emph{fuzzy interpretation of MultiLayer Perceptrons} (MLPs, \citealt{Haykin99}) obtained by encoding synaptic connections as weighted typicality inclusions \citep{JELIA2021}.
Then, the widespread interest in neural networks strongly motivates the development of proof methods for reasoning with weighted DL KBs.

Entailment for fuzzy DLs is in general undecidable \citep{CeramiStraccia2011, BorgwardtPenaloza12}, and this motivates the investigation of many-valued approximations of fuzzy multi-preferential entailment.
In particular, the finitely many-valued case is widely studied in the DL literature \citep{GarciaCerdanaAE2010,
BobilloDelgadoStraccia2012,BorgwardtPenaloza13}, and has been recently considered also in the context of weighted DL KBs \citep{ICLP22} by means of the notions of {\em coherent}, {\em faithful} and  {\em $\varphi$-coherent} models of such KBs, previously considered in the fuzzy case \citep{JELIA2021,ICLP22,ECSQARU2021}.
A proof-of-concept implementation in Answer Set Programming (ASP) and {\em asprin} \citep{BrewkaAAAI15} has been provided for the $\lc$ fragment of $\alc$, which is obtained by disabling roles, and universal and existential restrictions.
The approach adopts G\"odel connectives (or alternatively \L ukasiewicz connectives) and addresses $\varphi$-coherent entailment, a form of defeasible reasoning based on canonical \mbox{$\varphi$-coherent} models.
As concerns the complexity of the problem, a $\Pi^p_2$ upper bound was given \citep{ICLP22}, but the exact complexity is unknown.

This paper contributes to the understanding of the problem both from a theoretical point of view and on the practical side.
In fact, after introducing the required background (Section~\ref{sec:ALC}), the upper bound is improved to ${\sc P}^{\sc NP[log]}$ by showing an algorithm running in polynomial time and performing \emph{parallel} queries to an NP oracle ($\sc P^{||NP}$; Section~\ref{sec:upper-bound}).
As $\sc P^{||NP}$ is known to coincide with $\sc P^{NP[log]}$ \citep{BussHay91}, while $\Pi^p_2 = \sc P^{NP[log]}$ in unlikely to hold (unless the polynomial hierarchy collapses to $\sc P^{NP[log]}$), there must be space for improving the proof-of-concept implementation.
A contribution in this respect is given by the ASP encodings reported in Section~\ref{sec:encodings}, obtaining the desired multi-preferential semantics by taking advantage of weak constraints, possibly without the need for weights.
Further improvements at an asymptotic level are unlikely, as the problem is shown to be actually ${\sc P}^{\sc NP[log]}$-complete by giving a polynomial-time reduction of the \textsc{max sat odd} problem (\citealt{DBLP:journals/siamcomp/Wagner90}; Section~\ref{sec:lower-bound}), which amounts to determining whether the maximum number of jointly satisfiable clauses among a given set is an odd number.
Finally, the scalability of the different ASP encodings powering the implemented system is evaluated empirically on defeasible entailment queries over synthetic weighted DL KBs,
reporting results on 
KBs (Section~\ref{sec:experiment}) with large search spaces, while the earlier proof-of-concept implementation can only deal
with small KBs and search spaces.

\section{Weighted finitely-valued ${\lc}_n$ with typicality}  \label{sec:ALC}

Let ${\cal C}_n = \{0, \frac{1}{n},\ldots, \frac{n-1}{n}, \frac{n}{n}\}$, for an integer $n \geq 1$, denote the finitely-valued set of truth degree, also called \emph{truth space}.
The \emph{truth degree functions} $\otimes$, $\oplus$, $\ominus$ and $\rhd$ associated with the connectives $\wedge$, $\vee$, $\neg$ and $\rightarrow$, respectively, are the following:
$a \otimes b = min\{a,b\}$,  $a \oplus b = max\{a,b\}$, 
 $\ominus a = 1 - a$, and $a \rhd b= 1$ {\em if} $a \leq b$ {\em and} $b$ {\em otherwise}
(as in G\"odel logic with involutive negation). %
Let ${N_C}$ be a set of concept names and ${N_I}$ be a set  of individual names.  
The set  of ${\lc}_n$ \emph{concepts} is
defined inductively as follows:  
(i) $A \in N_C$, $\top$ and $\bot$ are {concepts};
(ii) if $C$ and $ D$ are concepts, then $C \sqcap D,\; C \sqcup D,\; \neg C$
are {concepts}.
An {\em ${\lc}_n$ KB}
$K$ is a pair $({\cal T}, {\cal A})$, where ${\cal T}$ (the TBox) is a set of {\em concept inclusions} of the form $C \sqsubseteq D \;\theta \alpha$,
and ${\cal A}$ (the ABox) is a set of {\em assertions} of the form $C(a) \; \theta\alpha$, 
with $C$ and $D$ being concepts, $a \in N_I$, $\theta \in \{\geq,\leq,>,<\}$  and $\alpha \in [0,1]$.
Concept inclusions and assertions are collectively called \emph{axioms}.

A {\em finitely many-valued interpretation} (short.\ interpretation) is a pair $I=\langle \Delta^I, \cdot^I \rangle$, where
$\Delta^I$ is a non-empty domain and 
$\cdot^I$ is an {\em interpretation function} that assigns 
to each $a \in N_I$ a value $a^I \in\Delta^I$, and
to each  $A\in N_C$ a function  $A^I :  \Delta^I \ri {\cal C}_n $.
Hence, a domain element $x \in \Delta^I$ belongs to the extension of a concept name $A \in N_C$ to some degree $A^I(x)$ in ${\cal C}_n$, and to a composed concept according to the following inductive definition:
\begin{align*}
    \begin{array}{rrr}
        \top^I(x) = 1 & \qquad (C \sqcap D)^I(x) = C^I(x) \otimes D^I(x) & \qquad (\neg C)^I(x) = \ominus C^I(x) \\
        \bot^I(x) = 0 & (C \sqcup D)^I(x) = C^I(x) \oplus D^I(x)
    \end{array}
\end{align*}

\noindent
The interpretation function $\cdot^I$ is also extended to axioms as follows:
\begin{align*}
    \begin{array}{ll}
        (C \sqsubseteq D)^I= \mathit{inf}_{x \in \Delta^I}  C^I(x) \rhd D^I(x) \qquad &
        (C(a))^I=C^I(a^I)
    \end{array}
\end{align*}
(note that in our setting the \emph{infimum truth degree} in  ${\cal C}_n$ in the above expression coincides with the \emph{minimum truth degree} in  ${\cal C}_n$).

\begin{definition}[Satisfiability and entailment for ${\lc}_n$ knowledge bases] \label{satisfiability}
Let $K=({\cal T}, {\cal A})$ be a weighted ${\lc}_n$ KB, and $I$ be an interpretation.
Relation $\models$ is defined as follows:
$I \models C \sqsubseteq D \; \theta \alpha$ if $(C \sqsubseteq D)^I \; \theta \alpha$;
%
$I \models C(a) \; \theta \alpha$ if $C^I(a^I) \; \theta \alpha$;
%
for a set $S$ of axioms, $I \models S$ if $I \models E$ for all $E \in S$;
%
$I \models K$ if $I \models {\cal T}$ and $I \models {\cal A}$.
If $I \models \Gamma$, we say that $I$ {\em satisfies} $\Gamma$ or that $I$ is a \emph{model} of $\Gamma$ (for $\Gamma$ being an axiom, a set of axioms, or a KB).
An axiom $E$ is \emph{entailed} by $K$, written $K \models E$, if $I \models E$ holds for all models $I$ of $K$.
\end{definition}


${\lc}_n$ is extended with typicality concepts of the form $\tip(C)$ so that the degree of membership of domain individuals in $C$ defines the typical elements of $C$. 
%
For an interpretation $I= \langle \Delta^I, \cdot^I \rangle$,
a preference relation $\prec_C$ on $\Delta^I$ (where $x \prec_C y$ means that {\em $x$ is preferred to $y$}) is obtained as follows:
for all $x, y \in \Delta^I$, $x \prec_C y$ if and only if $C^I(x) > C^I(y)$.
The typical elements of $C$ are those belonging to $C$ with the greatest positive truth degree. 
%
Formally, the interpretation of a typicality concept $\tip(C)$ is as follows: 
for all $x \in \Delta^I$, $(\tip(C))^I(x) = 0$ if there is $y \in \Delta^I$ such that $y \prec_C x$, and $C^I(x)$ otherwise.
When $(\tip(C))^I(x)>0$, $x$ is said to be a \emph{typical $C$-element} in $I$.
Note that each relation $\prec_C$ has  the properties of a preference relation in KLM-style ranked interpretations by \citealt{whatdoes}, that is,  $\prec_C$ is a modular and well-founded strict partial order.

A \emph{weighted typicality inclusion} has the form $\left(\tip(C) \sqsubseteq D, w\right)$, where $C$ and $D$ are concepts, and the weight $w$ is a real number;
concept $C$ is also said to be a \emph{distinguished concept}.
A \emph{weighted ${\lc}_n \tip$ KB} is a a tuple $\langle {\cal T}, {\cal D}, {\cal A} \rangle$, where the TBox $\mathcal{T}$ is a set of concept inclusions, ${\cal D}$  (defeasible TBox) is a set of weighted typicality inclusions, and ${\cal A}$ is a set of assertions.
For an interpretation $I = \langle \Delta^I, \cdot^I \rangle$, the \emph{weight of $x \in \Delta^I$ with respect to a distinguished concept $C$} is given by
$\mathit{weight}_C(x) = \sum_{\left(\tip(C) \sqsubseteq D, w\right) \in {\cal T}}{w \cdot D^I(x)}$.
Intuitively, the higher the value of $\mathit{weight}_C(x)$, the more typical is  $x$ relative to the defeasible properties of $C$.
The weight of an individual is then mapped to a truth degree by means of a monotonically non-decreasing function $\varphi: {\mathbb{R}} \rightarrow {\cal C}_n$, so that the notion of model can be naturally extended to weighted ${\lc}_n \tip$ KBs.
For example, the weighted  ${\lc}_n \tip$ KB $\langle \{\mathit{Tall} \sqcap \mathit{Small} \sqsubseteq  \bot \geq 1 \},$ $\{\mathit{\tip(Horse)} \sqsubseteq \mathit{Has\_Tail}, +50),$ $(\mathit{\tip(Horse)} \sqsubseteq \mathit{Tall},+40),$ $(\mathit{\tip(Horse)} \sqsubseteq   \mathit{Has\_Stripes}, -50)\}, \emptyset \rangle$ encodes that a horse normally has a tail and is tall, but usually does not have stripes. 
Accordingly, a tall horse with tail and without stripes is more typical than a tall horse with tail and stripes.
Moreover, as usual in preferential semantics, we restrict to canonical models, which are large enough to contain 
a domain element for any possible valuation of concepts. 

\begin{definition}[Canonical $\varphi$-coherent model and $\varphi$-coherent entailment]\label{varphi-coherence} 
Let $K=\langle {\cal T}, {\cal D}, {\cal A}  \rangle$ be a weighted ${\lc}_n \tip$ KB, and $\varphi: {\mathbb{R} } \rightarrow {\cal C}_n$ be a monotonically non-decreasing function.
An interpretation $I=\langle \Delta^I, \cdot^I \rangle$ is {\em $\varphi$-coherent} if  $C^I(x) = \varphi(\mathit{weight}_{C}(x))$ holds for each distinguished concept $C$ in ${\cal D}$ and for all $x \in \Delta^I$.
$I$ is a {\em $\varphi$-coherent model} of $K$ if it is a $\varphi$-coherent interpretation satisfying ${\cal T}$ and ${\cal A}$.
$I$ is a {\em canonical $\varphi$-coherent model} of $K$ if 
(i) $I$ is a $\varphi$-coherent model of $K$, and
(ii) for each $\varphi$-coherent model $J=(\Delta^J, \cdot^J)$ of $K$ and each $x \in \Delta^J$,
there is an element $y \in \Delta^I$ such that, for all concept names $A$ occurring in $K$, $A^I(y)=A^J(x)$.\footnote{
Note that the semantics adopted here slightly differs from the original definition given by \citet{ICLP22} in the interpretation of typicality concepts, which is not crisp in Definition~\ref{varphi-coherence}.
Anyway, the existence of canonical $\varphi$-coherent  models, for weighted KBs having at least a $\varphi$-coherent  model, can be proved as with the crisp interpretation of typicality concepts (see the supplementary material for paper \citealt{ICLP22}, Appendix A).}
An axiom $E$ is {\em $\varphi$-coherently entailed} by $K$ if $I \models E$ holds for all canonical $\varphi$-coherent models $I$ of $K$.
\end{definition}

According to the above definition, for every distinguished concept $C$, the degree of membership of typical $C$-elements is the same in all canonical $\varphi$-coherent models;
it is essentially the highest degree of membership among all $\varphi$-coherent models.
In the next sections, we take advantage of such a property to study $\varphi$-coherent entailment in the case in which typicality concepts only occur in ${\cal D}$ and in the query.
We prove that deciding $\varphi$-entailment of a query $\tip(C) \sqsubseteq D \;\theta \alpha$  is a $\sc{P^{NP[log]}}$-complete problem, 
we introduce several ASP encodings addressing the computational problem and investigate their scalability.

\section{Computing $\varphi$-coherent entailment in ASP is in $\sc{P^{NP[log]}}$ }\label{sec:upper-bound}

In this section we elaborate on the encoding by \citet{ICLP22} to obtain an upper bound on the complexity of deciding $\varphi$-coherent entailment of a typicality inclusion of the form $\tip(C_q) \sqsubseteq D_q \geq \alpha$ from a weighted ${\lc}_n \tip$ knowledge base $K=\langle {\cal T}, {\cal D}, {\cal A}  \rangle$.
Specifically, we first introduce a $\sc{P^{NP}}$ algorithm, and then refine it to obtain a $\sc{P^{NP[log]}}$ upper bound. 
To simplify the presentation, here we work under the assumption that all axioms in ${\cal T}$ and ${\cal A}$ use $\geq$ as their comparison operator, and in Section~\ref{sec:encodings} we will extend the result to the general case.

%
%
%
We associate with $K$ an ASP program $\Pi_K$ with the following main features:



\begin{itemize}

\item 
Names in $N_C$ and in $N_I$ occurring in $K$, as well as an $\mathit{anonymous}$ individual name, are encoded by constant terms (i.e., strings starting by lowercase), composed concepts such as $C \sqcap D$ are encoded by composed terms such as $\mathit{and}(c,d)$, and any $C \sqsubseteq D$ is encoded by $\mathit{impl}(c,d)$.
Predicates $\mathit{concept/1}$ and $\mathit{ind}/1$ are used to mark concepts and individual names in $K$, and each weighted typicality inclusion $(\tip(C) \sqsubseteq D, w)$ is encoded by the fact $\mathit{wti}(c,d,w)$.

\item 
${\cal C}_n$ is encoded by $\mathit{val}(0..n)$, and an interpretation $I$ is guessed by means of rules
\begin{align*}
    \{\mathit{eval}(c,X,V) : \mathit{val}(X)\} = 1 \leftarrow \mathit{ind}(X)
\end{align*}
for each $C \in N_C$,
so that an atom of the form $\mathit{eval}(c,x,v)$ means that $C^I(x) = \frac{v}{n} \in {\cal C}_n$.
Relation $\mathit{eval}/3$ is extended to complex concepts naturally.
Additionally, for any $C \sqsubseteq D$, the valuation $C^I(x) \rhd D^I(x)$ is obtained by the following rules:
\begin{align*}
    \mathit{eval}(\mathit{impl}(c,d),X,1) \leftarrow \mathit{eval}(c,X,V), \mathit{eval}(d,X,V'), V \leq V'.\\
    \mathit{eval}(\mathit{impl}(c,d),X,V') \leftarrow \mathit{eval}(c,X,V), \mathit{eval}(d,X,V'), V > V'.
\end{align*}

\item
Each concept inclusion $C \sqsubseteq D \geq \alpha$ in ${\cal T}$,
each assertion $C(a) \geq \alpha$ in the ABox ${\cal A}$, and 
each distinguished concept $C$ in ${\cal D}$ are enforced by the constraints
\begin{align*}
    & \bot \leftarrow \mathit{eval}(\mathit{impl}(c,d),X,V), V < \alpha.
    \qquad \bot \leftarrow \mathit{eval}(c,a,V), V < \alpha.\\
    & \bot \leftarrow \mathit{ind}(X), W = \mathit{\#sum}\{W_D * V_D, D : \mathit{wti}(c,D,W_D), \mathit{eval}(D,X,V_D)\}, \\
    & \phantom{\bot \leftarrow{}} \mathit{eval}(c,X,V),\ \mathit{valphi}(n,W,V'), V \neq V'.
\end{align*}
the last one imposing $\varphi$-coherence,
where $\mathit{valphi}/3$ is defined so that $\varphi(W) = \frac{V'}{n}$.
\end{itemize}
From Lemma~1 by \citet{ICLP22}, there is a duality relation between $\varphi$-coherent models $I=\la \Delta^I, \cdot^I\ra$ of $K$ and the answer sets of $\Pi_K$:
Let $C \in N_C$ and $\frac{v}{n} \in {\cal C}_n$.
If there is a $\varphi$-coherent model $I=\langle \Delta^I, \cdot^I \rangle $ for $K$ and $x \in \Delta^I$ such that $C^I(x)=\frac{v}{n}$, then there is an answer set $S$ of $\Pi_K$ such that $\mathit{eval}(c,\mathit{anonymous},v) \in S$, and vice-versa.

As for the query $\tip(C_q) \sqsubseteq D_q \geq \alpha$, entailment is decided by the following algorithm:
\begin{itemize}
\item[(a)]
 find the highest value $\frac{v}{n} \in {\cal C}_n$ such that there is a $\varphi$-coherent model $I$ of $K$ with $C_q^I(x)=\frac{v}{n}$ for some $x \in \Delta^I$;
\item[(b)]
verify that for all $\varphi$-coherent models $I$ of $K$ and all $x \in \Delta^I$, if $C_q^I(x)=\frac{v}{n}$ then $C_q^I(x) \rhd D_q^I(x) \geq \alpha$ holds
(note that the implication trivially holds when $v = 0$).
\end{itemize}

Step (a) identifies the degree of membership of typical $C_q$-elements (if any) by 
invoking multiple times an ASP solver for $\Pi_K$ extended with
\begin{align*}
    \bot \leftarrow \mathit{\#count}\{X : \mathit{ind}(X),\ \mathit{eval}(c_q,X,v)\} = 0
\end{align*}
in order to verify the existence of an answer set containing $\mathit{eval}(c_q,x,v)$, for some $x$. 
Specifically, the ASP solver is called no more than $n$ times, for decreasing $v = n, \ldots ,1$ and stopping at the first successful call; if none is successful, there are no typical $C$-elements, and the query is true.

Step (b) checks that, for the selected $\frac{v}{n}$, there is no answer set of $\Pi_K$ containing both $\mathit{eval}(c_q,x,v)$ and $\mathit{eval}(d_q,x,v')$ whenever $\frac{v}{n} \rhd \frac{v'}{n} < \alpha$, for any $x$.
It requires one additional call to the ASP solver to check that there is no answer set for $\Pi_K$ 
extended with
\begin{align*}
    & \mathit{counterexample} \leftarrow \mathit{eval}(c_q,X,v),\ \mathit{eval}(d_q,X,V'),\ V' > v,\ V' < \alpha. \\
    & \bot \leftarrow not \; \mathit{counterexample}.
\end{align*}

As the size of $\Pi_K$ and its extensions is polynomial in the size of $K$ and of the query, and no disjunctive head or recursive aggregate is used, each call to the ASP solver can be answered by a call to an NP oracle.
In the worst case, $n+1$ calls to the NP  oracle are performed, which gives $\sc{P^{NP}}$ upper bound on the complexity of the decision problem.

The upper bound can be refined by observing that step (a) can be executed in parallel for each $v = 1, \ldots, n$, and similarly step (b) can be speculatively checked for each value $\frac{v}{n}$, regardless from $\frac{v}{n}$ being the degree of membership of typical $C_q$-elements (if any).
Once the answers to such $2n$ calls are obtained, the decision problem can be answered by selecting the highest value $\frac{v}{n}$ for which calls of type (a) returned \emph{yes}, and returning the answer obtained for the corresponding call of type (b) --- all other answers to calls of type (b) are simply ignored.


\begin{theorem}[Strict complexity upper bound for $\varphi$-coherent entailment; restricted case]\label{thm:upper-bound-restricted}
Deciding $\varphi$-coherent entailment of a typicality inclusion  $\tip(C) \sqsubseteq D \geq \alpha$ from a weighted ${\lc}_n \tip$ KB $K$ requires a polynomial number of parallel queries to an NP-oracle, 
under the assumption that all axioms in $K$ use $\geq$ as their comparison operator.
\end{theorem}
It follows that the decision algorithm is in $\sc{P^{NP[log]}}$ \citep{BussHay91}.

\section{$\sc{P^{NP[log]}}$ lower bound for $\varphi$-coherent entailment}\label{sec:lower-bound}

A lower bound to the complexity of entailment is given in this section, actually holding already for a considerably restricted fragment of the language.

\begin{theorem}[Strict complexity lower bound for $\varphi$-coherent entailment]\label{thm:lower-bound}
Determine if a typicality inclusion $\tip(C) \sqsubseteq D \; \theta \alpha$
is $\varphi$-coherently entailed by a weighted  ${\lc}_n \tip$ KB $K = \langle {\cal T}, {\cal D}, {\cal A} \rangle$ is $\sc{P^{NP[log]}}$-hard, even if ${\cal T}$ and ${\cal A}$ are empty, $C$ and $D$ are concept names, and $\theta\alpha$ is fixed to $\geq 1$.
\end{theorem}

In the following, we provide a reduction from the problem \textsc{max sat even}, asking for the parity of the maximum number of jointly satisfiable clauses in a given set.
(The problem is often formulated as \textsc{max sat odd}, \citealt{DBLP:journals/siamcomp/Wagner90}).
Let $\Gamma = \{\gamma_1,\ldots,\gamma_n\}$ be a set of $n \geq 0$ clauses of propositional logic, and let $\mathit{vars}(\Gamma)$ be the set of boolean variables occurring in $\Gamma$.
We construct a weighted ${\lc}_n \tip$ KB $K_\Gamma = \langle \emptyset, {\cal D}_\Gamma, \emptyset \rangle$ and $\varphi : {\mathbb{R}} \rightarrow {\cal C}_n$ such that the maximum number of jointly satisfiable clauses is even if and only if $\tip(\mathit{Sat}) \sqsubseteq \mathit{Even}_n \geq 1$ is $\varphi$-coherently entailed by $K$.

Our construction uses
$\varphi(w) = \min(1, \max(0, \frac{w}{n}))$,
and 
${\cal D}_\Gamma$ comprising the following weighted typicality inclusions:
\begin{align}
    \label{eq:reduction:assignment}
    (\tip(A_x) \sqsubseteq A_x, n^2) & \qquad \forall x \in \mathit{vars}(\Gamma)\\
    \label{eq:reduction:clause:1}
    (\tip(C_i) \sqsubseteq \top, |\{x \mid \neg x \in \gamma_i\}| \cdot n) & \qquad \forall i = 1..n \\
    \label{eq:reduction:clause:2}
    (\tip(C_i) \sqsubseteq A_x, n) & \qquad \forall i = 1..n,\ \forall \phantom{\neg}x \in \gamma_i\\
    \label{eq:reduction:clause:3}
    (\tip(C_i) \sqsubseteq A_x, -n) & \qquad \forall i = 1..n,\ \forall \neg x \in \gamma_i\\
    \label{eq:reduction:sat}
    (\tip(\mathit{Sat}) \sqsubseteq C_i, 1) & \qquad \forall i = 1..n \\
    \label{eq:reduction:even:0}
    (\tip(\mathit{Even}_0) \sqsubseteq \top, n) \\
    \label{eq:reduction:even:i:1}    
    (\tip(\mathit{Even}_{i,1}) \sqsubseteq \mathit{Even}_{i-1}, -n)),
    (\tip(\mathit{Even}_{i,1}) \sqsubseteq \mathit{C}_i, n)) & \qquad \forall i = 1..n\\
    \label{eq:reduction:even:i:2}
    (\tip(\mathit{Even}_{i,2}) \sqsubseteq \mathit{Even}_{i-1}, n)),
    (\tip(\mathit{Even}_{i,2}) \sqsubseteq \mathit{C}_i, -n)) & \qquad \forall i = 1..n\\
    \label{eq:reduction:even:i}
    (\tip(\mathit{Even}_i) \sqsubseteq \mathit{Even}_{i,1}, n)),
    (\tip(\mathit{Even}_i) \sqsubseteq \mathit{Even}_{i,2}, n)) & \qquad \forall i = 1..n
%
 %
\end{align}
In a nutshell, \eqref{eq:reduction:assignment} enforces a crisp valuation for $A_x$, so that each $\varphi$-coherent interpretation $I = \langle \Delta^I, \cdot^I \rangle$ satisfying \eqref{eq:reduction:assignment} is one-to-one with a boolean assignment $I_\Gamma = \{x \mapsto A_x^I(y) \mid x \in \mathit{vars}(\Gamma)\}$ for $\Gamma$, where $y$ is any individual in $\Delta^I$;
\eqref{eq:reduction:clause:1}--\eqref{eq:reduction:clause:3} enforce $C_i^I(y) = I_\Gamma(\gamma_i)$;
\eqref{eq:reduction:sat} enforces $\mathit{Sat}^I(y) = \frac{k}{n}$, where $k = |\{i \mid i = 1..n,$ $I_\Gamma(\gamma_i) = 1\}|$;
\eqref{eq:reduction:even:0} enforces $\mathit{Even}_0^I(y) = 1$;
\eqref{eq:reduction:even:i:1}--\eqref{eq:reduction:even:i} enforce $\mathit{Even}_i^I(y) = \mathit{Even}_{i-1}^I(y) \text{ XOR } C_i^I(y)$.
All in all, $(\tip(\mathit{Sat}))^I(y) = \frac{k}{n} > 0$ if and only if $k$ is the maximum number of jointly satisfiable clauses in $\Gamma$, and $\mathit{Even}_n(y) = 1$ if and only if $k$ is even.
Therefore, the next result is established.

\begin{lemma}\label{lem:lower_bound}
There is a canonical $\varphi$-coherent model $I = \langle \Delta^I, \cdot^I \rangle$ of $K_\Gamma$ and an element $y \in \Delta$ such that $(\tip(\mathit{Sat}))^I(y) = \frac{k}{n} > 0$  and $\mathit{Even_n^I}(y)= 1$
if and only if $k$ is the maximum number of jointly satisfiable clauses in $\Gamma$ and $k$ is even.
\end{lemma}




Every canonical $\varphi$-coherent model $J$ of the knowledge base $K_\Gamma$ must contain a domain element with the same valuations as $y$ in  Lemma~\ref{lem:lower_bound} above.
Hence, in $J$ the membership degree of any domain element $z \in \Delta^I$ in the typicality concept $\tip(\mathit{Sat})$ is either $0$ (when $z$ is not a typical instance of $\mathit{Sat}$) or it is $\frac{k}{n}>0$, 
where $k$ is the maximum number of jointly satisfiable clauses.
Furthermore, $\mathit{Even_n^I}(z)= 1$, as the membership degree of $z$ in $\mathit{Even_n^I}$ only depends on $k$.
Hence, the next result is established, and Thorem~\ref{thm:lower-bound} proved.

\begin{proposition}
The entailment $K_\Gamma \models \tip(Sat) \sqsubseteq Even_n \geq 1$ holds if and only if the maximum number $k$ of jointly satisfiable clauses in $\Gamma$ is even.
\end{proposition}

\section{Comparing different ASP encodings of $\varphi$-coherent entailment}\label{sec:encodings}

We present four ASP encodings improving the one in Section~\ref{sec:upper-bound} both in terms of generality and of scalability.
The encodings adopt a combination of several ASP constructs, among them \mbox{@-terms}, custom propagators, weak constraints and weight constraints.
First of all, the input is encoded by the following facts (with weights represented as integers):
\begin{itemize}
\item
\lstinline|valphi($v$,$\mathit{LB}$,$\mathit{UB}$)| whenever $\varphi(w) = \frac{v}{n}$ if and only if $\mathit{LB} < w \leq \mathit{UB}$ holds;

\item
\lstinline|concept($C$)| for each relevant concept $C$, where named concepts are represented as constant terms, and complex terms by means of the uninterpreted functions \lstinline|and|, \lstinline|or|, \lstinline|neg| and \lstinline|impl|;

\item
\lstinline|ind($a$)| for each individual name $a$, among them the \lstinline|anonymous| one;

\item
\lstinline|concept_inclusion($C$,$D$,$\theta$,$\alpha$)| for each concept inclusion $C \sqsubseteq D \;\theta \frac{\alpha}{n}$;

\item
\lstinline|assertion($C$,$a$,$\theta$,$\alpha$)| for each assertion $C(a) \;\theta \frac{\alpha}{n}$;

\item
\lstinline|wti($C$,$D$,$w$)| for each weighted typicality inclusion $(\tip(C) \sqsubseteq D, w)$;

\item
\lstinline|query($C_q$,$D_q$,$\theta$,$\alpha$)| for the typicality inclusion $\tip(C_q) \sqsubseteq D_q \;\theta \frac{\alpha}{n}$;

\item
\lstinline|crisp($C$)| as an optimization for $(\tip(C) \sqsubseteq C, \infty)$, to enforce a crisp evaluation of concept $C$ (where $\infty$ is a sufficiently large integer to obtain $\varphi(\infty \cdot \frac{1}{n}) = 1$; see equation~(1) for an example);

\item
\lstinline|exactly_one($\mathit{ID}$)| and \lstinline|exactly_one_element($\mathit{ID}$,$C_i$)| ($i = 1..k$)
to optimize 
$\top \sqsubseteq C_1 \sqcup \cdots \sqcup C_k \geq 1$ (at least one)
and
$C_i \sqcap C_j \geq 1$ with $j = i+1..k$ (at most one);
%
\end{itemize}
%
%
%
The latter two predicates are useful to express membership of individuals in mutually exclusive concepts.
Moreover, the following interpreted functions are implemented via @-terms:
\lstinline|@is_named_concept($C$)|, returning $1$ if $C$ is a named concept, and $0$ otherwise;
%
\lstinline|@min($v$,$v'$)|,
\lstinline|@max($v$,$v'$)|,
\lstinline|@neg($v$)|, and
\lstinline|@impl($v$,$v'$,$n$)|,
for the truth degree functions $\otimes$, $\oplus$, $\ominus$ and $\rhd$ in G\"odel logic (other truth degree functions can be considered, see Section~\ref{sec:ALC}).

\begin{figure}
\figrule
\begin{lstlisting}[basicstyle=\scriptsize\ttfamily,numberblanklines=false]
/*\label{ln:base:first}*/val(0..n).   concept(bot).   eval(bot,X,0) :- ind(X).   concept(top).   eval(top,X,n) :- ind(X).

/*\label{ln:base:guess:first}*/{eval(C,X,V) : val(V)} = 1 :- concept(C), ind(X), @is_named_concept(C) = 1, not crisp(C).
/*\label{ln:base:guess:last}*/{eval(C,X,0); eval(C,X,n)} = 1 :- concept(C), ind(X), @is_named_concept(C) = 1, crisp(C).

/*\label{ln:base:eval:first}*/eval(and(A,B),X,@min(V,V'))  :- concept(and(A,B)), eval(A,X,V), eval(B,X,V').
eval( or(A,B),X,@max(V,V'))  :- concept( or(A,B)), eval(A,X,V), eval(B,X,V').
eval(neg(A),X,@neg(V)) :- concept(neg(A)),   eval(A,X,V).
eval(impl(A,B),X,@impl(V,V',n)) :- concept(impl(A,B)), eval(A,X,V), eval(B,X,V').
/*\label{ln:base:eval:last}*/:- concept(C), @is_named_concept(C)!=1, crisp(C); ind(X), not eval(C,X,0), not eval(C,X,n).

/*\label{ln:base:axioms:first}*/:- concept_inclusion(C,D,$\theta_>$,$\alpha$), eval(impl(C,D),X,V), not V $\theta_>$ $\alpha$.
/*\label{ln:base:axioms:individual}*/ind(ci(C,D,$\theta_<$,$\alpha$)) :- concept_inclusion(C,D,$\theta_<$,$\alpha$).
:- concept_inclusion(C,D,$\theta_<$,$\alpha$), eval(impl(C,D),ci(C,D,$\theta_<$,$\alpha$),V), not V $\theta_<$ $\alpha$.
/*\label{ln:base:axioms:last}*/:- assertion(C,X,$\theta$,$\alpha$); eval(C,X,V), not V $\theta$ $\alpha$.

/*\label{ln:base:exactly-one}*/:- exactly_one(ID), ind(X), #count{C : exactly_one_element(ID,C), eval(C,X,n)} != 1.

% find the largest truth degree for the left-hand-side concept of query 
/*\label{ln:base:query:left}*/:~ query(C,_,_,_), eval(C,X,V), V > 0. [-1@V+1]

% verify if there is a counterexample to the truth of query ($\theta_<$) or to its falsity ($\theta_>$)
/*\label{ln:base:query:right:first}*/typical(C,X) :- query(C,_,_,_), eval(C,X,V), V = #max{V' : eval(C,X',V')}.
witness :- query(C,D,$\theta_>$,$\alpha$); typical(C,X), eval(impl(C,D),X,V), not V $\theta_>$ $\alpha$.
witness :- query(C,D,$\theta_<$,$\alpha$); typical(C,X), eval(impl(C,D),X,V), $\phantom{\mathtt{not}\,}$ V $\theta_<$ $\alpha$.
/*\label{ln:base:query:right:last}*/:~ witness. [-1@1] 

/*\label{ln:base:show:first}*/#show witness : witness.
/*\label{ln:base:show:last}*/#show eval(C,X,V) : witness, eval(C,X,V), concept(C), @is_named_concept(C) = 1.
\end{lstlisting}
\caption{Base encoding, with $\theta \in \{\geq,\leq,>,<\}$, $\theta_> \in \{>, \geq\}$, and $\theta_< \in \{<, \leq\}$}\label{fig:enc:base}
\figrule
\end{figure}





The \emph{base encoding} is shown in Figure~\ref{fig:enc:base}.
Line~\ref{ln:base:first} introduces the truth degrees from ${\cal C}_n$ and fixes the interpretation of $\bot$ and $\top$.
Lines~\ref{ln:base:guess:first}--\ref{ln:base:guess:last} guess a truth degree for named concept, using only crisp truth degrees for crisp concepts.
Lines~\ref{ln:base:eval:first}--\ref{ln:base:eval:last} evaluate composed concepts, and impose crisp truth degrees for crisp concepts.
Lines~\ref{ln:base:axioms:first}--\ref{ln:base:axioms:last} enforce concept inclusions and assertions;
note that, by the semantic definition given in Section~\ref{sec:ALC}, concept inclusions with $\geq$ and $>$ define properties holding for all individuals, while concept inclusions with $\leq$ and $<$ define properties holding for at least one individual; such an individual is introduced by line~\ref{ln:base:axioms:individual}.
Line~\ref{ln:base:exactly-one} enforces \emph{exactly one} constraints.
Line~\ref{ln:base:query:left} expresses a preference for assigning a large truth degree to $C_q$.
Lines~\ref{ln:base:query:right:first}--\ref{ln:base:query:right:last} define typical $C_q$-elements and express a weaker preference for the existence of a witness:
if the query uses $\theta_> \in \{>,\geq\}$, a witness is a $\varphi$-coherent model $I$ and an element $x \in \Delta^I$ such that $(\tip(C_q))^I \rhd D_q^I(x) \;\theta_> \frac{\alpha}{n}$ holds (i.e., $x$ makes the query false), and the query is true if such a witness does not exist;
if the query uses $\theta_< \in \{<,\leq\}$, a witness is a $\varphi$-coherent model $I$ and an element $x \in \Delta^I$ such that $(\tip(C_q))^I \rhd D_q^I(x) \;\theta_< \frac{\alpha}{n}$ holds (i.e., $x$ makes the query true), and the query is false if such a witness does not exist.
Lines~\ref{ln:base:show:first}--\ref{ln:base:show:last} report in the output whether a witness was found (and the truth degrees it assigns to named concepts).

The encoding must be enriched with the enforcement of
$\varphi$-coherence.
A first solution is the addition, for each distinguished 
concept $C$, of a \emph{custom propagator} that infers \mbox{\lstinline|eval($C$,$x$,$v$)|} whenever $\varphi(\mathit{weight}_C(x)) = \frac{v}{n}$.
In case of conflict, the propagator provides
\begin{align*}
    \lstinline|:- eval($D_1$,$x$,$v_1$), ..., eval($D_k$,$x$,$v_k$), not eval($C$,$x$,$v$).|
\end{align*}
as the reason of inference, where $(\tip(C) \sqsubseteq D_i,w_i)$, for $i = 1..k$, are all the weighted typicality inclusions for $C$ in ${\cal T}$ and \lstinline|eval($D_i$,$x$,$v_i$)| is true in the current assignment.

\begin{figure}
\figrule
\begin{lstlisting}[basicstyle=\scriptsize\ttfamily]
/*\label{ln:order:query:left}*/:~ query(C,_,_,_), eval_ge(C,X,V). [-1@2]

/*\label{ln:order:space:first}*/{eval_ge(C,X,V) : val(V), V > 0} :- concept(C), ind(X).
/*\label{ln:order:space:last}*/:- eval_ge(C,X,V), V > 1, not eval_ge(C,X,V-1).  % $C^I(x) \geq \frac{v}{n} \Longrightarrow C^I(x) \geq \frac{v-1}{n}$

% $C^I(x) = \frac{v}{n} \Leftrightarrow C^I(x) \geq \frac{v}{n} \text{ and } C^I(x) < \frac{v+1}{n}$
/*\label{ln:order:map:first}*/:- concept(C), ind(X); eval(C,X,V), V > 0; not eval_ge(C,X,V).
:- concept(C), ind(X); eval(C,X,V); eval_ge(C,X,V+1).
/*\label{ln:order:map:last}*/:- concept(C), ind(X); eval_ge(C,X,V), not eval_ge(C,X,V+1); not eval(C,X,V).

% $(A \sqcap B)^I(x) \geq \frac{v}{n} \Longleftrightarrow A^I(x) \geq \frac{v}{n} \text{ and } B^I(x) \geq \frac{v}{n}$ 
/*\label{ln:order:complex:first}*/:- concept(and(A,B)), ind(X), eval_ge(and(A,B),X,V); not eval_ge(A,X,V).
:- concept(and(A,B)), ind(X); eval_ge(and(A,B),X,V); not eval_ge(B,X,V).
:- concept(and(A,B)), ind(X); eval_ge(A,X,V), eval_ge(B,X,V); not eval_ge(and(A,B),X,V).

% $(A \sqcup B)^I(x) \geq \frac{v}{n} \Longleftrightarrow A^I(x) \geq \frac{v}{n} \text{ or } B^I(x) \geq \frac{v}{n}$ 
:- concept(or(A,B)), ind(X); eval_ge(or(A,B),X,V); not eval_ge(A,X,V), not eval_ge(B,X,V).
:- concept(or(A,B)), ind(X); eval_ge(A,X,V); not eval_ge(or(A,B),X,V).
:- concept(or(A,B)), ind(X); eval_ge(B,X,V); not eval_ge(or(A,B),X,V).

% $(\neg A)^I(x) \geq \frac{v}{n} \Longleftrightarrow A^I(x) \leq 1 - \frac{v}{n}$ 
:- concept(neg(A)), ind(X); eval_ge(neg(A),X,V); eval_ge(A,X,n-V+1).
:- concept(neg(A)), ind(X), val(V), V > 0; not eval_ge(A,X,n-V+1); not eval_ge(neg(A),X,V).

% $(A \sqsubseteq B)^I(x) \geq \frac{v}{n} \Longleftrightarrow A^I(x) \leq B^I(x) \text{ or } B^I(x) \geq \frac{v}{n}$ 
l_gt_r(A,B,X) :- concept(impl(A,B)), ind(X); eval_ge(A,X,V); not eval_ge(B,X,V).
:- concept(impl(A,B)), ind(X); eval_ge(impl(A,B),X,V); l_gt_r(A,B,X); not eval_ge(B,X,V).
:- concept(impl(A,B)), ind(X), val(V), V>0; not l_gt_r(A,B,X); not eval_ge(impl(A,B),X,V).
/*\label{ln:order:complex:last}*/:- concept(impl(A,B)), ind(X); eval_ge(B,X,V); not eval_ge(impl(A,B),X,V).
\end{lstlisting}
\caption{Rules replacing line~\ref{ln:base:query:left} of the base encoding to obtain the order encoding}\label{fig:enc:order}
\figrule
\end{figure}

The base encoding is not suitable to obtain a strict upper bound for our problem, due to the weak constraint in line~\ref{ln:base:query:left} using a linear number of levels with respect to the size of ${\cal C}_n$.
Such levels can be removed by replacing \lstinline|[-1@V+1]| with 
\lstinline|[-2$^V$@1]| \citep{DBLP:journals/ia/Alviano18}, 
which however results into a weighted preference relation
giving a $\sc P^{NP}$ upper bound \citep{DBLP:conf/lpnmr/BuccafurriLR97}.
Removing such weights is less trivial, nonetheless can be achieved by introducing atoms representing $C^I(x) \geq \frac{v}{n}$, that is, an \emph{order encoding} 
\citep{DBLP:journals/constraints/TamuraTKB09}
for finitely many-valued interpretations as shown in Figure~\ref{fig:enc:order}.
Note that the level in line~\ref{ln:order:query:left} can be removed by replacing \lstinline|[-1@2]| with \lstinline|[-2@1]|, and in turn the weight \lstinline|-2| can be removed by introducing two copies of the weak constraint using \lstinline|[-1@1, 1]| and \mbox{\lstinline|[-1@1, 2]|}.
As for the other rules, lines~\ref{ln:order:space:first}--\ref{ln:order:space:last} define the search space for predicate \lstinline|eval_ge/2|,
lines~\ref{ln:order:map:first}--\ref{ln:order:map:last} match the assignment of \lstinline|eval/2| and \lstinline|eval_ge/2|, and
lines~\ref{ln:order:complex:first}--\ref{ln:order:complex:last} implement inferences of \lstinline|eval_ge/2| over composed concepts.
We therefore have a generalization of Theorem~\ref{thm:upper-bound-restricted}.

\begin{theorem}[Strict complexity upper bound for $\varphi$-coherent entailment]\label{thm:upper-bound-general}
Deciding $\varphi$-coherent entailment of a typicality inclusion  $\tip(C) \sqsubseteq D \;\theta \alpha$ from a weighted ${\lc}_n \tip$ KB $K$ is in $\sc{P^{NP[log]}}$.
\end{theorem}

\begin{figure}
\figrule
\begin{lstlisting}[basicstyle=\scriptsize\ttfamily]
/*\label{ln:wc:base:first}*/% $C^I(x) = \frac{v}{n} \Longleftrightarrow \mathit{LB} < \mathit{weight}_C(x) \leq \mathit{UB}$
:- val(V), val_phi(V,LB,UB); wti(C,_,_), ind(X); eval(C,X,V);
   not LB < #sum{W*VD, D,VD : wti(C,D,W), eval(D,X,VD)} <= UB.
:- val(V), val_phi(V,LB,UB); wti(C,_,_), ind(X); not eval(C,X,V);
/*\label{ln:wc:base:last}*/   LB < #sum{W*VD, D,VD : wti(C,D,W), eval(D,X,VD)} <= UB.

/*\label{ln:wc:order:first}*/% $C^I(x) \geq \frac{v}{n} \Longleftrightarrow \mathit{weight}_C(x) > \mathit{LB}$
:- val(V), V > 0, val_phi(V,LB,UB); wti(C,_,_), ind(X); eval(C,X,V);
   #sum{W ,D,VD : wti(C,D,W), eval_ge(D,X,VD)} > LB.
:- val(V), V > 0, val_phi(V,LB,UB); wti(C,_,_), ind(X); not eval(C,X,V);
/*\label{ln:wc:order:last}*/   #sum{W, D,VD : wti(C,D,W), eval_ge(D,X,VD)} > LB.
\end{lstlisting}
\caption{Rules added to the base encoding (lines~\ref{ln:wc:base:first}--\ref{ln:wc:base:last}) and to the order encoding (lines~\ref{ln:wc:order:first}--\ref{ln:wc:order:last}) to enforce $\varphi$-coherence via weight constraints.}\label{fig:enc:wc}
\figrule
\end{figure}

 
Even if the custom propagators provide a sensible performance gain with respect to the previously implemented encoding, indeed settling the grounding bottleneck, they miss the opportunity for several deterministic and inexpensive inferences.
An alternative way to enforce $\varphi$-coherence
is given by the weight constraints reported in Figure~\ref{fig:enc:wc}, for both the base and order encodings, leading to the results in Section~\ref{sec:experiment}.
The idea is to just check membership of $\mathit{weight}_C(x)$ in the intervals of interest, without materializing its actual value so to avoid the reintroduction of the grounding bottleneck.

\section{Experiment}\label{sec:experiment}

The encoding by \cite{ICLP22} was shown to work as a proof-of-concept for small instances, and even the variation described in Section~\ref{sec:upper-bound} is already challenged by KBs corresponding to fully-connected neural networks with 20 \emph{binary} inputs and 150 weighted typicality inclusions.
The size of the search space is around $10^6$, since, for such KBs, it is given by the combination of values for concepts corresponding to input nodes, as in fact the values of the other nodes is implied.
We observed that the number of weighted typicality inclusions has a significant impact on the size of the grounding of these encodings.

We therefore focus on the encodings presented in Section~\ref{sec:encodings}, and consider synthetic KBs encoding fully-connected neural networks with one input layer, two hidden layers and one output node;
nodes are encoded by concept names, edges are encoded by weighted typicality inclusions, and there are edges from any node in a layer to any node in the next layer.
We consider KBs of four different dimensions, comprising 50/100/200/400 nodes, including 10/20/40/80 input nodes.
For each dimension, we generated 10 instances by randomizing edge weights.
As for the query, we fix it to $\tip(O) \sqsubseteq I_1 \sqcup I_2 \geq 0.5$, asking whether truth of one of the first two input nodes $I_1, I_2$ implies truth of the output node $O$, with a truth confidence of $0.5$.

\begin{figure}
    \figrule
    \begin{tikzpicture}[scale=.55]
        \pgfkeys{/pgf/number format/set thousands separator = {}}
        \begin{axis}[
            draw=black,
            text=black,
            width=0.6\textwidth,
            height=0.6\textwidth,
            legend style={at={(1.5,1)}, anchor=north, align=left},
            legend cell align=left,
            reverse legend,
            xlabel={\textsc{base encoding}},
            ylabel={\textsc{order encoding}},
            xmin=0,
            xmax=1800,
            ymin=0,
            ymax=1800,
            xtick={0,300,600,900,1200,1500,1800},
            ytick={0,300,600,900,1200,1500,1800},
            tick label style={/pgf/number format/1000 sep=\,},
            grid=both,
            title={\normalsize{$|{\cal C}_n| = 5$}},
        ]
            \draw [dashed, red] (rel axis cs:0,0) -- (rel axis cs:1,1);

            \addplot [mark size=3pt, only marks, color=green!50!black, mark=triangle] table[col sep=tab, skip first n=1, x index=4, y index=8] {scatter-5.csv};
            \addlegendentry{\,$80/400/38240$}
            \addplot [mark size=3pt, only marks, color=orange, mark=+] table[col sep=tab, skip first n=1, x index=3, y index=7] {scatter-5.csv};
            \addlegendentry{\,$40/200/\phantom{0}9520$}
            \addplot [mark size=3pt, only marks, color=blue, mark=o] table[col sep=tab, skip first n=1, x index=2, y index=6] {scatter-5.csv};
            \addlegendentry{\,$20/100/\phantom{0}2360$}
            \addplot [mark size=3pt, only marks, color=purple, mark=x] table[col sep=tab, skip first n=1, x index=1, y index=5] {scatter-5.csv};
            \addlegendentry{\,$10/\phantom{0}50/\phantom{00}580$}
        \end{axis}
    \end{tikzpicture}
    $\phantom{xxxx}$
    \begin{tikzpicture}[scale=.55]
        \pgfkeys{/pgf/number format/set thousands separator = {}}
        \begin{axis}[
            draw=black,
            text=black,
            width=0.6\textwidth,
            height=0.6\textwidth,
            xlabel={\textsc{base encoding}},
            ylabel={\textsc{order encoding}},
            xmin=0,
            xmax=1800,
            ymin=0,
            ymax=1800,
            xtick={0,300,600,900,1200,1500,1800},
            ytick={0,300,600,900,1200,1500,1800},
            tick label style={/pgf/number format/1000 sep=\,},
            grid=both,
            title={\normalsize{$|{\cal C}_n| = 10$}},
        ]
            \draw [dashed, red] (rel axis cs:0,0) -- (rel axis cs:1,1);

            \addplot [mark size=3pt, only marks, color=green!50!black, mark=triangle] table[col sep=tab, skip first n=1, x index=4, y index=8] {scatter-10.csv};
            \addlegendentry{$80/400/38240$}
            \addplot [mark size=3pt, only marks, color=orange, mark=+] table[col sep=tab, skip first n=1, x index=3, y index=7] {scatter-10.csv};
            \addlegendentry{$40/200/\phantom{0}9520$}
            \addplot [mark size=3pt, only marks, color=blue, mark=o] table[col sep=tab, skip first n=1, x index=2, y index=6] {scatter-10.csv};
            \addlegendentry{$20/100/\phantom{0}2360$}
            \addplot [mark size=3pt, only marks, color=purple, mark=x] table[col sep=tab, skip first n=1, x index=1, y index=5] {scatter-10.csv};
            \addlegendentry{$10/\phantom{0}50/\phantom{00}580$}
            
            \legend{}
        \end{axis}
    \end{tikzpicture}

    \bigskip
    {
        \begin{minipage}{\textwidth}
            \scriptsize
            \tabcolsep=0.125cm
            \begin{tabular}{crrrcrrrcrrr}
                \toprule
                &&&& \multicolumn{4}{c}{$|{\cal C}_n| = 5$} & \multicolumn{4}{c}{$|{\cal C}_n| = 10$}\\
                \cmidrule{5-8}
                \cmidrule{9-12}
                & \multicolumn{3}{c}{Size (number of)} && \multicolumn{3}{c}{Runtime (seconds)} && \multicolumn{3}{c}{Runtime (seconds)} \\
                \cmidrule{2-4}
                \cmidrule{6-8}
                \cmidrule{10-12}
                & inputs & nodes & edges & Solved &  min  & avg & max & Solved &  min  & avg & max \\
                \cmidrule{1-12}
                \parbox[t]{2mm}{\multirow{4}{*}{\rotatebox[origin=c]{90}{\textsc{order}}}} &
                10 & 50 & 580 & 
                90\% & 4 & 139 & 798 &
                40\% & 6 & 393	&  1534
                \\
                & 20 & 100 & 2360 & 
                60\% & 15 &	30	& 70 &
                50\% & 21	& 24	& 30
                \\
                & 40 & 200 & 9520 & 
                70\% & 67 & 79 &118 &
                50\% & 94 & 242 & 766
                \\
                & 80 & 400 & 38240 &
                60\% & 298	& 309	& 339 &
                50\% & 400 & 412 & 433
                \\
                \cmidrule{1-12}
                \parbox[t]{2mm}{\multirow{4}{*}{\rotatebox[origin=c]{90}{\textsc{base}}}} &
                10 & 50 & 580 &
                40\% & 4 & 465 & 1639 &
                20\% & 6 & 7 & 8 
                \\
                & 20 & 100 & 2360 & 
                50\% & 16 & 21 & 34 &
                50\% & 22 & 89 & 150
                \\
                & 40 & 200 & 9520 & 
                70\% & 69 & 96 & 187 &
                60\% & 95 & 180 & 444
                \\
                & 80 & 400 & 38240 &
                60\% & 415 & 608 & 1125 &
                40\% & 500 & 813 & 1330
                \\
                \bottomrule
            \end{tabular}
        \end{minipage}
    }
    \caption{
        Runtime (in seconds) of the base and order encodings relying on weight constraints to answer queries over weighted KBs encoding fully-connected neural networks of different dimensions (10 for each dimension).
        The KBs have a concept for each node, and a weighted typicality inclusion for each edge.
        In the scatter plots, timeouts are normalized to 1800 seconds.
    }\label{fig:experiment}
    \figrule
\end{figure}

The experiment was run on an Intel Xeon 5520 2.26 GHz, with
runtime limited to 30 minutes.
Figure~\ref{fig:experiment} reports data on running times for answering the queries using the truth spaces ${\cal C}_4$ and ${\cal C}_9$, that is, 5 and 10 truth degrees;
the resulting search spaces have sizes ranging from $5^{10}$ (around $10^7$) to $10^{80}$.
Data is reported for the base and order encodings relying on the use of weight constraints; the results using the custom propagator are worse.
The percentage of 10 cases solved within a timeout of 30 minutes is shown, as well as the minimum, average and maximum time for the solved instances.
The two scatter plots highlight that, with a few exceptions, the order encoding provides a performance gain to the system.
Finally, there is an impact of the number of truth degrees on performance, but there could be space for a compromise between the level of approximation of reasoning and the consumed computational resources.

\section{Related Work}

Fuzzy description logics (DLs) have been widely studied in the literature for representing vagueness in DLs, e.g., by 
\cite{Stoilos05}, \cite{LukasiewiczStraccia09}, \cite{BorgwardtPenaloza12}, 
based on the idea that concepts and roles can be interpreted as fuzzy sets and fuzzy relations.
In fuzzy DLs, formulas have a truth degree from a truth space  $\cal S$, usually either the interval $[0, 1]$, as  in Mathematical Fuzzy Logic \citep{Cintula2011},
or the finitely-valued set ${\cal C}_n$.
Moreover, \emph{truth degree functions} $\otimes$, $\oplus$, $\ominus$ and $\rhd$ are associated with the connectives $\wedge$, $\vee$, $\neg$ and $\rightarrow$, respectively, and often chosen as t-norm, s-norm, implication function and negation function in some well known system of many-valued logic \citep{Gottwald2001}.
%
The finitely-valued case is well studied for DLs \citep{GarciaCerdanaAE2010,
BobilloDelgadoStraccia2012,BorgwardtPenaloza13}, and in this paper we have considered a variant of the boolean fragment ${\lc}_n$ of the finitely-valued 
$\alc$ with typicality considered by \cite{ICLP22}, by considering a \emph{many-valued interpretation of typicality concepts} rather than a crisp one.
We have taken ${\cal C}_n$ as the truth degree set and we have restricted our consideration to G\"odel  many-valued logic with involutive negation. 

${\lc}_n$ is extended with typicality concepts of the form $\tip(C)$, in the spirit of the extension of $\alc$ with typicality in the two-valued case \citep{FI09}, but taking into account that in the many-valued case the degree of membership of domain individuals in a concept $C$ is used to identify the typical elements of $C$. 
While the semantics has strong relations with  KLM logics by \cite{KrausLehmannMagidor:90} and with other preferential semantics, such as {\em c-representations} \citep{Kern-Isberner01} which also consider weights, we have adopted a {\em concept-wise} multi-preferential semantics, in which different  preferences $\prec_C$ are associated with different concepts $C$. This also makes our formalism different form the one considered by \cite{CasiniStracciaLPAR2013}, in their  rational closure construction for fuzzy logic.
The choice of a many-valued interpretation of the  typicality operator has been first considered by \cite{AGT_arXiv_Argumentation2022}
to develop a conditional semantics for gradual argumentation.

Finally, the weighted KBs considered in this paper also relates to works dealing with preference combination. Among them the preferred subtheories and the ranked KBs introduced by \cite{Brewka89,Brewka04},  the algebraic framework for preference combination in multi-relational contextual hierarchies proposed by \cite{BozzatoEK21},
and the work on concept combination based on typicality by \cite{Lieto2018}.

%


\section{Conclusions} 

Defeasible reasoning over weighted ${\cal LC}_n$ KBs is a computationally intensive task, previously addressed in the finitely many-valued case by adopting solving techniques suitable for problems in the complexity class $\Pi^p_2$ \citep{ICLP22}.
As shown in Section~\ref{sec:upper-bound}, the ASP encoding powering the available solution in the literature can be the basis for defining an algorithm asking all required queries to the NP oracle in parallel, and then inspecting the obtained answers to decide if the entailment holds.
We therefore refined the upper bound on the complexity of the problem to $\sc P^{||NP} = P^{NP[log]}$, which we also proved to be optimal as the problem is also $\sc{P^{NP[log]}}$-hard (Section~\ref{sec:lower-bound}).

On a more practical side, in Section~\ref{sec:encodings} we revised the previously proposed ASP encoding by taking advantage of several linguistic extensions and coding techniques for ASP, among them @-terms, custom propagators, weak constraints, weight constraints
and order encoding.
While all such constructs improve readability of the code, it turns out that the implementation and maintenance of the custom propagator has a higher cost than the others.
In fact, the implemented custom propagator was very helpful to settle the grounding bottleneck, but it was also clear that capturing all deterministic and inexpensive inferences was nontrivial.
A pondered use of weight constraints showed to be more rewarding, performing better on the verification of typicality properties of the test cases considered in Section~\ref{sec:experiment}.
Source code is available at \url{https://github.com/alviano/valphi}.

A natural direction to extend this work is by introducing more flexibility on the activation function, giving to the user the possibility to use different $\varphi_i$ functions for different concepts $C_i$.
This is in fact a semantic extension already considered by \cite{AGT_arXiv_Argumentation2022}, and it would enable the application of our system to the verification of typicality properties of MultiLayer Perceptrons (MLPs) with different activation functions for different layers.
This work is also a step towards the definition of proof methods for reasoning from weighted KBs under a finitely many-valued 
preferential semantics 
 in more expressive and lightweight DLs, under different many-valued logics, 
 as for the ${\cal EL}$ case  \citep{JELIA2021}.





\end{document}